\documentclass[preprints,article,accept,pdftex,moreauthors]{Definitions/mdpi} 
\usepackage{tikz}
\usetikzlibrary{
    shapes.geometric, 
    arrows.meta,      
    positioning,      
    calc              
}
\usepackage{graphicx}
\firstpage{1} 
\pubvolume{1}
\issuenum{1}
\articlenumber{0}
\pubyear{2025}
\copyrightyear{2025}
\datereceived{ } 
\daterevised{ } 
\dateaccepted{ } 
\datepublished{ } 
\hreflink{https://doi.org/} 

\Title{Open-Vocabulary Object Detection in UAV Imagery: A Review and Future Perspectives}

\TitleCitation{Title}

\Author{Yang Zhou$^{1}$, Junjie Li$^{1}$, CongYang Ou$^{1}$,Dawei Yan$^{1}$, Haokui Zhang$^{1,*}$, Xizhe Xue$^{1,*}$}

\address{%
$^{1}$ \quad Northwestern Polytechnical University\\
$^{*}$ \quad Corresponding Author}

\abstract{Due to its extensive applications, aerial image object detection has long been a hot topic in computer vision. In recent years, advancements in Unmanned Aerial Vehicles (UAV) technology have further propelled this field to new heights, giving rise to a broader range of application requirements. However, traditional UAV aerial object detection methods primarily focus on detecting predefined categories, which significantly limits their applicability. The advent of cross-modal text-image alignment (e.g., CLIP) has overcome this limitation, enabling open-vocabulary object detection (OVOD), which can identify previously unseen objects through natural language descriptions. This breakthrough significantly enhances the intelligence and autonomy of UAVs in aerial scene understanding. This paper presents a comprehensive survey of OVOD in the context of UAV aerial scenes.
We begin by aligning the core principles of OVOD with the unique characteristics of UAV vision, setting the stage for a specialized discussion. Building on this foundation, we construct a systematic taxonomy that categorizes existing OVOD methods for aerial imagery and provides a comprehensive overview of the relevant datasets. This structured review enables us to critically dissect the key challenges and open problems at the intersection of these fields. Finally, based on this analysis, we outline promising future research directions and application prospects. This survey aims to provide a clear road map and a valuable reference for both newcomers and seasoned researchers, fostering innovation in this rapidly evolving domain. We keep tracing related works at https://github.com/zhouyang2002/OVOD-in-UVA-imagery
}


\begin{document}


\section{Introduction}

In recent years, unmanned aerial vehicle (UAV) technology has achieved dramatic development, with corresponding application scenarios expanding quickly. UAVs, commonly known as drones, have evolved from early entertainment devices into an indispensable tools~\cite{javed2024state,mao2024survey}. Their application scenarios have extend from precision agriculture~\cite{sadgrove2018real} and infrastructure inspection to public safety~\cite{chaturvedi2021machine,fang2024strategies}and disaster response~\cite{albahri2024systematic,reilly2010detection}. 

The significant impact of UAVs stems primarily from their unique aerial perspective, which enables the collection of extensive visual data and provides comprehensive environmental awareness. To fully utilize aerial imagery, UAVs must possess autonomous perception and environmental interpretation capabilities~\cite{wang2020uav,nelson2019view}. As a fundamental computer vision task, object detection serves as the core technology enabling this intelligent perception. The development of deep learning has led to major breakthroughs in this field. Advanced object detection algorithms based on architectures like the YOLO series~\cite{redmon2017yolo9000,redmon2016you} and Faster R-CNN~\cite{ren2016faster} have achieved remarkable results. When applied to UAV-captured scenes, these models can efficiently and accurately identify and locate target objects, supporting various intelligent applications ranging from automated wildlife monitoring~\cite{yi2021ai} to real-time traffic analysis~\cite{zhao2003car}.

However, the efficacy of these traditional methods is fundamentally constrained by their "closed-set" design~\cite{closedset,vaze2021open}. They are trained on large-scale, meticulously annotated datasets and can only recognize a predefined, fixed set of object categories. This inherent limitation creates a significant bottleneck for UAV operating in the real world, which is often dynamic, unstructured, and unpredictable. For instance, a UAV tasked with post-earthquake assessment cannot be pretrained to recognize every possible type of debris or sign of human activity. The prohibitive cost of annotating exhaustive datasets for every potential scenario, coupled with the inability to adapt to novel objects, severely curtails the autonomy and practical utility of UAV systems.

\begin{figure}
\includegraphics[width= \textwidth]{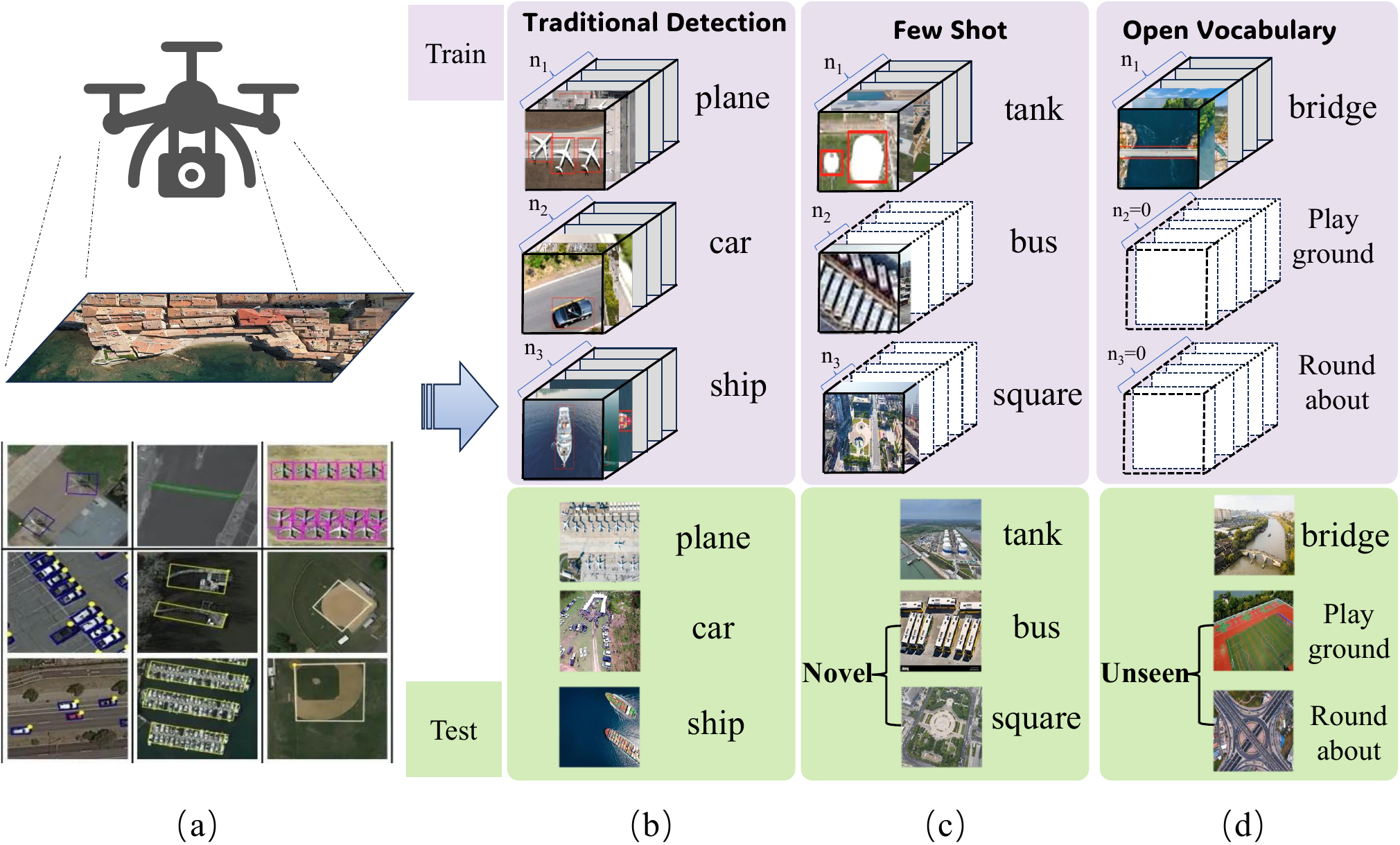}
\caption{Comparison between traditional close-set detection, few shot detection and open vocabulary detection. (a) UAV image object detection; (b) Traditional close-set detection. All test categories have appeared in the training set; (c) Few shot detection. Part test categories have only a few training samples; (d) Open vocabulary detection. Some categories have never been seen in the training set, only their names are provided.}
\label{fig: comparison}
\end{figure} 

To address the limitations of traditional closed-set recognition, researchers have turned to a new approach called Open-Vocabulary Object Detection (OVOD) ~\cite{zhao2017openvocabularysceneparsing,zareian2021openvocabularyobjectdetectionusing,gu2022openvocabularyobjectdetectionvision}. OVOD represents a fundamental change in object detection, shifting from "recognizing only predefined categories" to "detecting any object that can be described in language." This advancement has been made possible by recent developments in Vision-Language Models (VLMs), particularly the CLIP model ~\cite{clip}. The key innovation of OVOD lies in using the rich semantic knowledge stored in these models. Unlike traditional methods that rely on fixed category labels, OVOD models learn to match visual features from image regions with corresponding text descriptions. In practice, users can simply input new text queries, and the model will automatically identify image regions that match these descriptions. This approach gives the detector strong capabilities to recognize new objects, even ones it has never seen during training. A simple comparison between traditional closed-set detection, few-shot detection, and open-vocabulary detection is illustrated in Figure \ref{fig: comparison}.

The combination of OVOD and drone technology is especially powerful, offering new possibilities for smarter and more flexible operations. The ability to find objects using simple text commands is particularly useful when we don't know exactly what we're looking for in advance. Here are some practical examples:

\begin{itemize}
\item Emergency Rescue: After disasters, rescue teams can tell drones to look for important things like "people needing help", "temporary shelters", "red cross signs", or "damaged buildings". With OVOD, the drone can understand these new instructions immediately without extra training.

\item Wildlife Protection: Conservationists can use drones to watch large natural areas for problems like hunting or logging. They can give commands like "find hunter camps", "look for hurt elephants", or "spot illegal cutting tools", even if these things are not in the original training data.

\item Smart Cities: City officials can use drones with OVOD to check urban areas for problems. The system can understand commands like "find cars parked illegally near fire hydrants", "look for big road holes", or "find fallen branches blocking sidewalks".
\end{itemize}

In simple terms, OVOD turns regular drones into smart helpers that can understand what people need and find things accordingly. This makes them much more useful in real-world situations where needs can change quickly. While combining OVOD and drone technology shows great promise, this research area is still new but growing fast. A systematic review is crucial to summarize existing methods, identify key challenges, and guide future research. This paper aims to provide the first comprehensive survey on this topic. The main contributions of our work are as follows:






\begin{itemize}

\item \textbf{Theoretical Integration:} We thoroughly examine the core concepts of OVOD and align them with the specific requirements and limitations of UAV-based aerial vision.

\item \textbf{Comparative Analysis:} We introduce a structured taxonomy to organize existing OVOD techniques, offering a clear framework for analyzing their architectures and learning mechanisms. Finally, we provide an concise comparison between different types of existing methods. 

\item \textbf{Challenge Analysis:} We investigate major obstacles and unresolved issues in aerial OVOD, including domain shifts, object scale variations, and the absence of specialized benchmarks.

\item \textbf{Research Directions:} We highlight promising future trends and potential breakthrough applications to motivate and shape further advancements in this field.

\end{itemize}

The remainder of this survey is organized as follows: \hyperref[sec:back]{Section 2} discusses the foundational background and related work in both traditional and open-vocabulary detection. \hyperref[sec:method]{Section 3} summarizes the exiting methods into two major types and conducts concise introduction and comparison. \hyperref[sec:dataset]{Section 4} reviews the datasets and evaluation metrics pertinent to the field. \hyperref[sec:challenge]{Section 5} provides a deep dive into the prevalent challenges and open issues. Finally, \hyperref[sec:future]{Section 6} concludes the survey by offering a perspective on future research directions.

  
\section{Background}
\label{sec:back}

In this section, we introduce the basic concepts related to this survey. First, we look at standard object detection methods used with drone images and discuss their main limitations. Next, we explain the fundamentals of OVOD, including its core ideas and key supporting technologies. Finally, we explore the special challenges of drone-captured images that require customized OVOD approaches.

\subsection{Traditional Object Detection}
Object detection in aerial images captured by UAV has been a hot topic for over a decade. Early approaches relied on hand-crafted features, but the field was revolutionized by the advent of Deep Convolution Neural Networks (DCNNs). Modern UAV object detection is dominated by deep learning-based methods, which can be broadly categorized into two families~\cite{one-and-two-stage}: two-stage detectors and one-stage detectors.

Two-stage detectors approach the detection task using a region proposal and refined pipeline. This architecture first generates a sparse set of class-agnostic candidate object region proposals, or Regions of Interest (RoI), from the input image. In the second stage, it extracts features for each RoI and performs fine-grained classification and bounding box regression to produce the final detections. This methodology, while often computationally intensive, is renowned for its high accuracy and precise object localization.
The evolution of this family is marked by several seminal works:
\begin{itemize}
\item R-CNN~\cite{R-CNN} was the progenitor of this approach, pioneering the use of a deep convolution network for feature extraction on region proposals generated by an external algorithm like Selective Search. While groundbreaking, its multi-stage training and slow inference limited its practical application.
\item Fast R-CNN~\cite{fast-rcnn} significantly improved speed and efficiency by sharing the convolution feature computation across all proposals on an image. It introduced the RoIPool layer to extract fixed-size feature maps from variable-sized RoIs, enabling an end-to-end training process for the classification stage.
\item Faster R-CNN~\cite{faster-rcnn} represented a major leap by integrating the proposal generation step directly into the network. It introduced the Region Proposal Network (RPN), a fully-convolution network that shares features with the detection network, allowing for nearly cost-free, data-driven region proposals and enabling a unified, end-to-end trainable detection framework. Due to its robustness and high accuracy, Faster R-CNN became a dominant baseline for numerous aerial object detection tasks.
\item Mask R-CNN~\cite{mask-rcnn} extended Faster R-CNN by adding a parallel branch for predicting a pixel-level object mask in addition to the bounding box. It also introduced the RoIAlign layer, which replaced RoIPool to more precisely align the extracted features with the input, leading to significant gains in both detection and instance segmentation accuracy.
\end{itemize}

In contrast to the two-stage paradigm, one-stage detectors eschew the explicit region proposal step and instead treat object detection as a direct regression and classification problem. These models predict bounding boxes and class probabilities simultaneously in a single pass over the image, typically by dividing the image into a grid and having each cell predict potential objects. This unified architecture grants them a significant advantage in inference speed, making them highly suitable for real-time applications, a critical requirement for onboard processing on UAV.
Key innovations within the one-stage family have progressively addressed their initial accuracy limitations, particularly for the challenging small objects common in aerial scenes:
\begin{itemize}
\item YOLO~\cite{yolov1,redmon2017yolo9000,redmon2018yolov3} was the first model to frame object detection as a single regression problem, directly predicting bounding box coordinates and class probabilities from full images in one evaluation. Its architecture enabled unprecedented real-time performance, fundamentally shifting the research landscape.
\item SSD~\cite{SSD} introduced the concept of using multiple feature maps from different layers of a network to detect objects at various scales. By making predictions from both deep (low-resolution) and shallow (high-resolution) feature maps, SSD significantly improved the detection of small objects, a pervasive challenge in UAV imagery.
\item RetinaNet~\cite{retinanet} identified the extreme imbalance between the foreground and background classes during training as a primary obstacle to the precision of a detector of one stage. It introduced the Focal Loss~\cite{FocalLoss}, a novel loss function that dynamically adjusts the cross-entropy loss to down-weight the contribution of easy, well-classified examples, thereby focusing the training on a sparse set of hard-to-detect objects.
\end{itemize}

To address the challenges of aerial scenes, researchers have introduced numerous modifications to these baseline architectures. For instance, to combat the prevalence of small objects, techniques like Feature Pyramid Networks (FPN)~\cite{fpn} were widely adopted to create multi-scale feature representations, ensuring that fine-grained details from small objects are preserved. Specialized anchor generation strategies and data augmentation techniques, such as copy-pasting small objects onto various backgrounds, were also employed~\cite{smallobject}. To handle class imbalance and complex backgrounds, loss functions like Focal Loss~\cite{focalimprove} were integrated to down-weight the contribution of easily classified negative examples. 

Despite these advancements, a fundamental and pervasive limitation underpins nearly all traditional object detectors: the Closed-Set Assumption~\cite{geng2020recent,yang2020convolutional}. These models~\cite{yolov5,yolov8-impr,varghese2024yolov8} are designed to operate within a "closed world," where the set of object categories is predefined and fixed at training time. The final classification layer of the network has a fixed number of output neurons, each corresponding to a specific class seen during training (e.g., 'car', 'person', 'building'). Consequently, these systems are inherently incapable of recognizing objects belonging to novel unseen categories. If a new object class (e.g., "a dropped parcel" or "a rare species of animal") needs to be detected, the entire pipeline, from data collection and annotation to model retraining and deployment, must be repeated. This process is not only resource-intensive and time-consuming, but also renders the system brittle and nonadaptive in dynamic, real-world scenarios where unforeseen objects are the norm, not the exception. This closed-set constraint is the primary bottleneck that motivates the exploration of open-vocabulary detection for more flexible and scalable aerial intelligence.

\begin{figure}
\includegraphics[width= \textwidth]{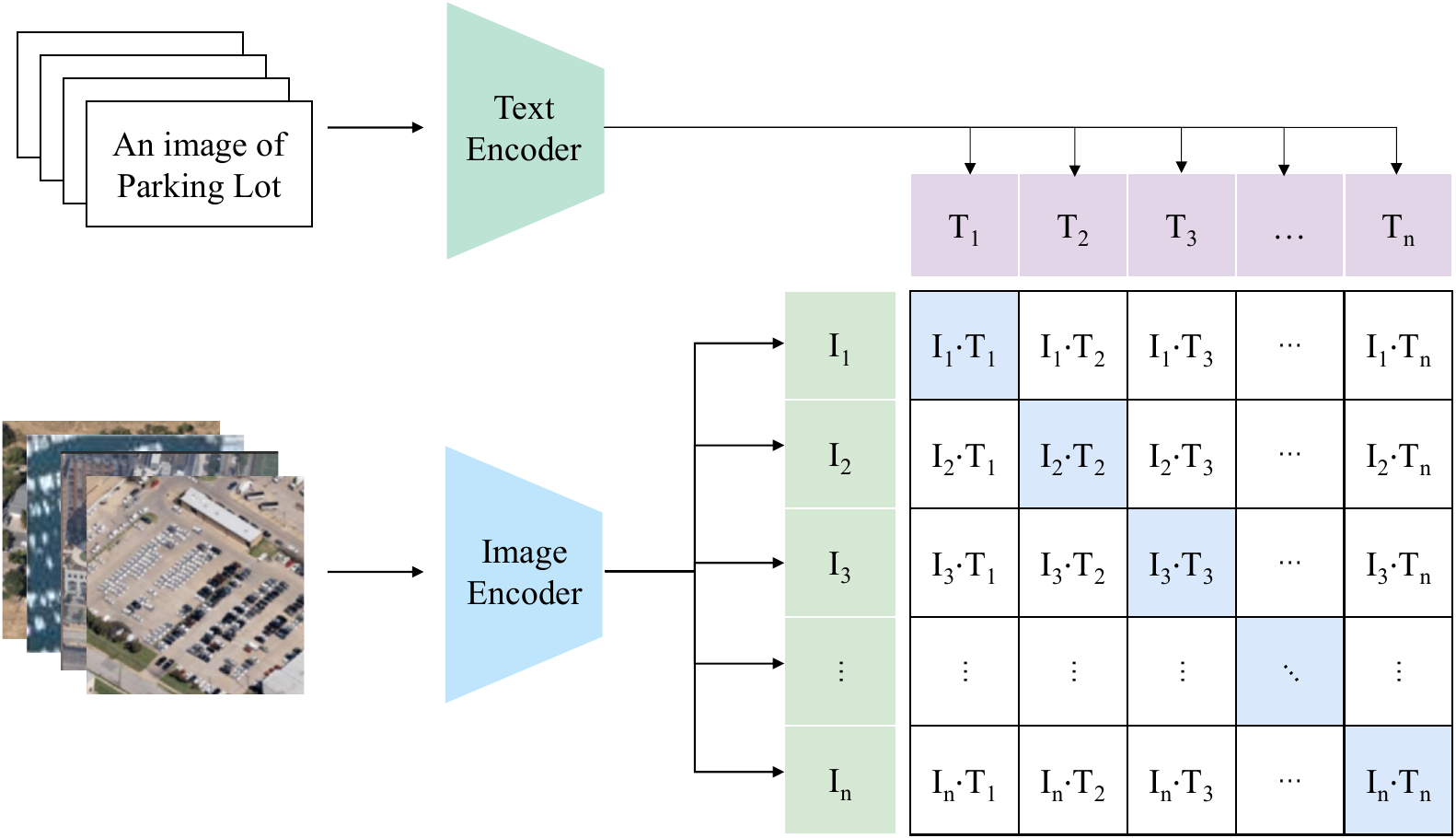}
\caption{Illustration of how CLIP is used in open vocabulary object detection. Region features extracted from the image are matched against category embeddings generated from textual prompts, enabling category expansion and unsupervised detection.}
\label{fig: clip}
\end{figure} 

\subsection{Fundamentals of Open-Vocabulary Object Detection}
OVOD emerges as a revolutionary paradigm designed to dismantle the barriers of the closed-set assumption. It empowers a detector to identify and localize objects described by arbitrary, open-ended textual queries, including categories never seen during the model's training phase~\cite{gu2021open,minderer2022simple,zareian2021openvocabularyobjectdetectionusing,du2022learning}.
The core idea of OVOD is to reframe object detection from a task of classification over a fixed set of categories to a task of region-text alignment. Instead of learning to map visual features to a specific class index, an OVOD model learns to map visual features of an object to the semantic representation of its corresponding textual description in a shared embedding space.
We illustrate the difference between open vocabulary object detection and traditional object detection and few shot in Figure \ref{fig: comparison}, mainly from the perspective of training and testing sets. Among them, the few shot method requires a small number of samples to train when encountering novel classes, while open vocabulary object detection does not require any at all.

The advanced capabilities of Open Vocabulary Object Detection (OVOD) systems stem from cross-modal alignment technologies, particularly visual language models such as CLIP. CLIP employs an elegant yet highly effective training approach. The model is trained on an extensive dataset containing 400 million image-text pairs collected from the internet. Architecturally, CLIP consists of two core components: a visual encoder that processes image inputs and a text encoder that processes corresponding natural language descriptions. During its training phase, the model is presented with a batch of these image-text pairs. The visual encoder generates an embedding for each image, while the text encoder does the same for each corresponding text caption. The core learning mechanism is driven by a contrastive objective:
Positive Pairs: For each correctly matched image and its corresponding text caption in the batch, the model is trained to make their vector embeddings as similar as possible. This teaches the model to associate an image with its authentic description.
Negative Pairs: In contrast, for every incorrectly matched pairing within the batch, that is, a given image paired with any other text caption from that same batch, the model is trained to make their embeddings as dissimilar as possible. This forces the model to learn fine-grained distinctions and not just general concepts. Through this process, repeated on a colossal scale, CLIP learns a rich shared embedding space where visual concepts and semantic meanings are aligned. An image of a dog and the text "a photo of a dog" will be positioned very close to each other in this space, while the text "a photo of a cat" will be pushed far away. This powerful and generalized alignment is the key that enables the model to understand and detect novel object categories described purely by text.

For OVOD, CLIP provides the "engine" for the alignment step, as shown in Figure \ref{fig: clip}. OVOD architectures leverage CLIP's pretrained encoders or fine-tune them to ensure that the visual features from image regions and the semantic features from arbitrary text prompts are comparable within this meaningful shared space~\cite{rasheed2022bridginggapobjectimagelevel,zhong2021regionclipregionbasedlanguageimagepretraining}. This pretrained knowledge of the visual world, linked to natural language, is precisely what allows an OVOD detector to recognize a "solar panel" or a "swimming pool" without ever having been explicitly trained on annotated bounding boxes for those classes. 

\subsection{Unique Characteristics of UAV Imagery}
While OVOD presents a powerful general framework, its direct and naive application to UAV imagery is fraught with challenges. The unique properties of aerial scenes can severely degrade the performance of models pre trained on ground level, generic web data. Understanding these characteristics is the first step toward developing robust UAV-centric OVOD solutions.
\begin{itemize}
\item \textbf{Small Objects:} Perhaps the most frequently cited challenge in aerial imaging. Due to high flight altitudes and wide-angle lenses, objects of interest (e.g., people, vehicles) often occupy a minuscule portion of the image, sometimes spanning only a few dozen or even a handful of pixels. For traditional detectors, this leads to the loss of distinguishing features after successive down sampling operations in deep neural networks. For OVOD, the challenge is even more nuanced: the low-resolution visual features extracted from such small objects may be too coarse and ambiguous to align reliably with a rich textual description in the VLM's embedding space. The visual signature of a 10x10 pixel "person" is weak and could be easily confused with other small, vertical structures, making a confident visual-linguistic match difficult.

\item \textbf{High Density and Cluttered Backgrounds:} UAV platforms are often used to monitor crowded scenes like parking lots, public squares, or disaster sites. This results in images containing a high density of objects, often with significant occlusion. Differentiating between individual instances in a dense crowd or a packed car park is a severe test for any detector's localization capabilities. Furthermore, aerial scenes are rife with complex and cluttered backgrounds—rooftop paraphernalia, varied terrain, foliage, and urban infrastructure. For an OVOD system, this clutter introduces a high number of "distractors." The visual features of background elements (e.g., a complex pattern of shadows, an air conditioning unit) might accidentally have a high similarity score with one of the text prompts, leading to false positives.

\item \textbf{Varying Viewpoints:} The vast majority of images used to train VLMs like CLIP are ground level, depicting objects from a frontal or near-frontal perspective. However, UAV primarily capture imagery from nadir (top-down) or oblique viewpoints. This creates a significant domain gap. The visual appearance of a "car" or a "person" from directly above is drastically different from its appearance in a typical photograph. An OVOD model relying on CLIP's pretrained knowledge may struggle because its internal concept of "car" is strongly tied to side views, not roof views. This viewpoint disparity can weaken the visual-linguistic alignment, which is the cornerstone of OVOD's success.

\item \textbf{Varying Scales:} The operational nature of UAVs, which can dynamically change their flight altitude, introduces extreme variations in object scale—often within the same mission or video sequence. An object that appears large when the UAV is flying low can become a tiny speck as it ascends. This multi-scale challenge requires a detector to be robustly invariant to scale. For OVOD, this means the visual encoder must produce consistent embeddings for the same object category across a wide range of resolutions, a non-trivial requirement that pushes the limits of standard VLM backbones.

\item \textbf{Challenging Imaging Conditions:} UAV operations are not confined to perfect, sunny days. The quality of captured imagery can be significantly degraded by a variety of factors. Illumination changes (e.g., harsh sunlight creating deep shadows, or low-light conditions at dawn/dusk) can obscure object details. Adverse weather like rain, fog, or haze can reduce contrast and introduce artifacts. Finally, motion blur, caused by the UAV's movement or a low shutter speed, can smear features. Each of these factors introduces noise and corrupts the input to the visual encoder, weakening the feature representations and making the subsequent alignment with clean text embeddings less reliable and more prone to error.
\end{itemize}
In summary, this section has established the limitations of traditional detection methods, introduced the promising paradigm of OVOD, and critically outlined the domain-specific challenges of UAV imagery. The confluence of these factors defines the core research problem: How can we adapt and advance Open Vocabulary Object Detection to perform robustly and accurately despite the severe challenges posed by small objects, high density, viewpoint and scale variations, and difficult imaging conditions in aerial scenes? The following chapters will survey the emerging body of work that seeks to answer this question.

\section{Methodology: OVOD for UAV Aerial Image}
\label{sec:method}


The rapid development of OVOD has led to many different methods. While early methods for ground view images could be easily grouped [1-5], drone images create special challenges that need custom solutions. To help organize these methods, this section divides OVOD approaches made for aerial images. According to how they learn to recognize new objects, we divide current methods into two major types: 1) Pseudo-Labeling Methods.  These help when there aren't many labeled drone images. They use a teacher-student system to create labels for unlabeled images, which helps the model learn better; 2) CLIP-driven integrations methods. This is the most common approach. It adds CLIP's ability to recognize objects without training (zero shot) into different object detectors.

\begin{figure}
\includegraphics[width = \textwidth]{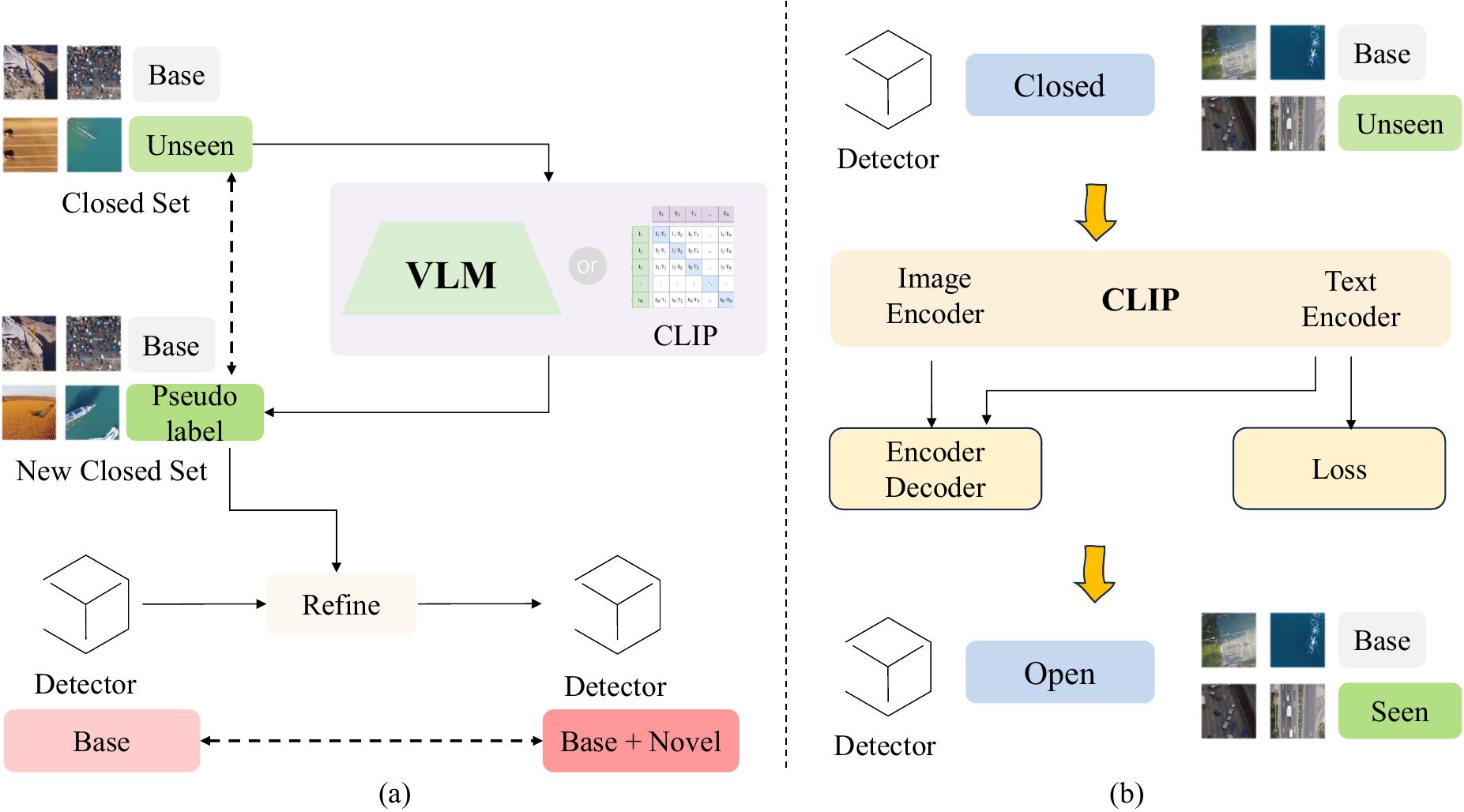}
\caption{Comparison of open-vocabulary object detection strategies. (a) Pseudo-Labeling Based Methods. (b) CLIP-driven Integration Methods.}
\label{fig:tech}
\end{figure}   

\subsection{Pseudo-Labeling Based Methods}
This class of methods addresses the critical bottleneck of limited annotated data in the aerial domain. As illustrated in Figure \ref{fig:tech})(a), the core principle is to use a powerful, pretrained model to generate "pseudo-labels" for vast quantities of unlabeled images. These pseudo-labeled data are then used to train or fine tune a primary object detector, effectively expanding its knowledge base beyond the initial, small set of labeled examples. This paradigm is particularly relevant for UAV applications, where acquiring unlabeled video footage is often trivial, but expert annotation is prohibitively expensive.


A pioneering example in this category is CastDet~\cite{li2024toward}. which introduces a CLIP-enhanced teacher-student framework specifically designed for aerial OVOD. The system consists of three core components: a student model (the primary detector), a localization teacher (a stabilized version of the student that generates reliable object proposals), and an external teacher (the fixed domain-specific vision language model RemoteCLIP ~\cite{liu2024remoteclipvisionlanguagefoundation}). The framework operates through an iterative self-training process: the localization teacher first identifies potential object regions in unlabeled images, which are then labeled by the external teacher. These temporary annotations train the student model to recognize novel classes, and the student's improved knowledge is subsequently transferred back to the localization teacher, creating a continuous self-improvement cycle termed the "flywheel effect." To maintain training stability, CastDet employs a dynamic label queue and a proposal selection strategy based on regression stability analysis, ensuring only high-quality candidate regions are used for learning.

Another highly relevant methodology, presented in ~\cite{saini2025advancing}, pushes this concept further towards open-set discovery. Instead of just detecting predefined novel classes, this work proposes a pipeline to discover and assign semantic labels to entirely unknown objects. It first employs a standard closed-set detector to identify both known objects and background regions, which are treated as potential unknown objects. These unknown regions are then fed into a powerful MLLM with a descriptive prompt, such as "What are the structured objects in this image?". The MLLM generates rich and natural language descriptions, from which novel class nouns (e.g., "sand volleyball court", "industrial building") are extracted using NLP techniques~\cite{anil2023palm2technicalreport,touvron2023llama2openfoundation}. Finally, a domain-adapted vision language model like RemoteCLIP~\cite{liu2024remoteclipvisionlanguagefoundation} is used to refine and validate these discovered labels by computing the image-text similarity.

The primary strength of these methods is their exceptional data efficiency. They significantly reduce the reliance on manual annotation and expand the detector's vocabulary. However, their performance is heavily dependent on the quality of the generated pseudo-labels. This can lead to error propagation, where incorrect pseudo-labels from the teacher model can mislead the student, degrading performance. Furthermore, multi-stage pipelines, as seen in the MLLM-based discovery approach, often suffer from high latency, making them more suitable for offline data annotation and analysis rather than real-time UAV deployment.

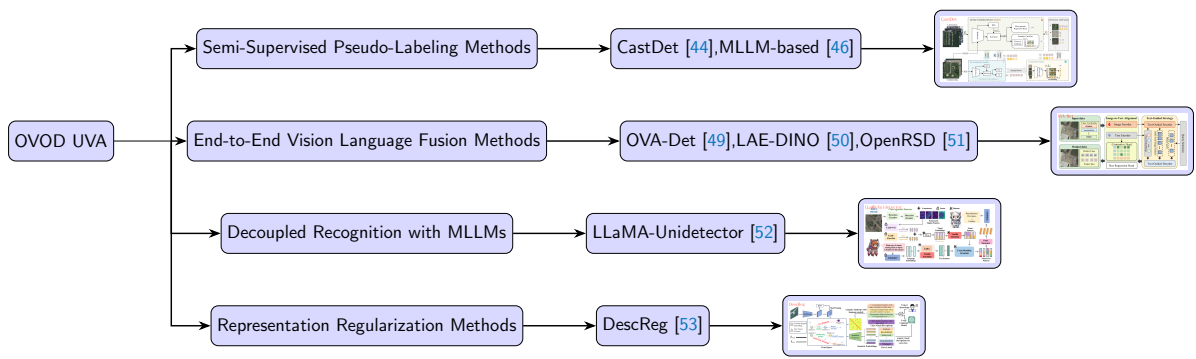
\begin{figure}
    \centering
    \resizebox{\textwidth}{!}{%
        \begin{tikzpicture}[%
            node distance = 1.0cm and 1.4cm,
            mybox/.style = {
                rectangle, 
                rounded corners, 
                draw=black, 
                fill=blue!15,          
                align=center,          
                minimum height=2.2em,  
                font=\sffamily         
            },
            myarrow/.style = {
                -Stealth, 
                thick
            }
        ]
        \node[mybox] (root) {OVOD UVA};
        \node[mybox, right=of root] (kd) {End-to-End Vision Language Fusion Methods};
        \node[mybox, above=of kd]   (so) {Semi-Supervised Pseudo-Labeling Methods};
        \node[mybox, below=of kd]   (di) {Decoupled Recognition with MLLMs};
        \node[mybox, below=of di]   (er) {Representation Regularization Methods};
        \node[mybox, right=of so] (sf)  {CastDet~\cite{li2024toward},MLLM-based~\cite{saini2025advancing}};
        \node[mybox, right=of kd] (wk) {OVA-Det~\cite{wei2024ova},LAE-DINO~\cite{pan2025locate},OpenRSD~\cite{huang2025openrsd}};
        \node[mybox, right=of di] (sl) {LLaMA-Unidetector~\cite{LLaMA-Unidetector}};
        \node[mybox, right=of er] (sm) {DescReg~\cite{zang2024descReg}};
        \node[mybox, right=of sf] (sfimg) {\includegraphics[width=2.5cm]{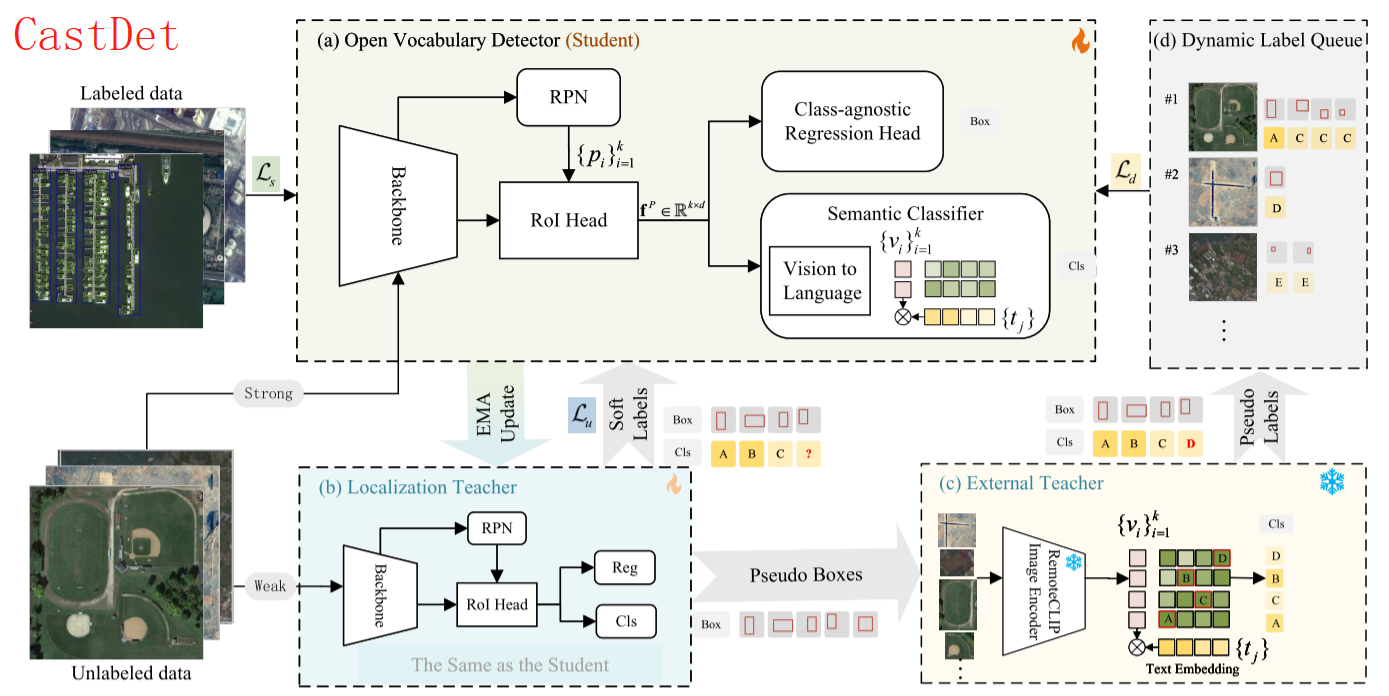}};
        \node[mybox, right=of wk] (wkimg) {\includegraphics[width=2.5cm]{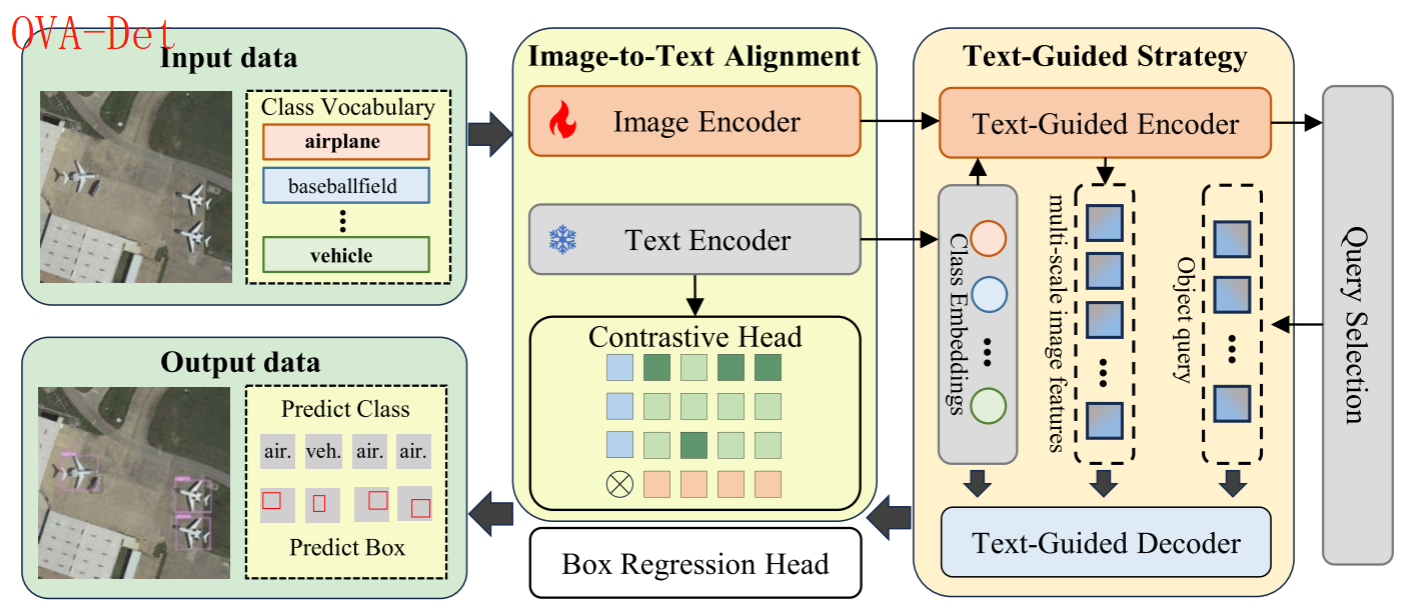}};
        \node[mybox, right=of sl] (slimg) {\includegraphics[width=2.5cm]{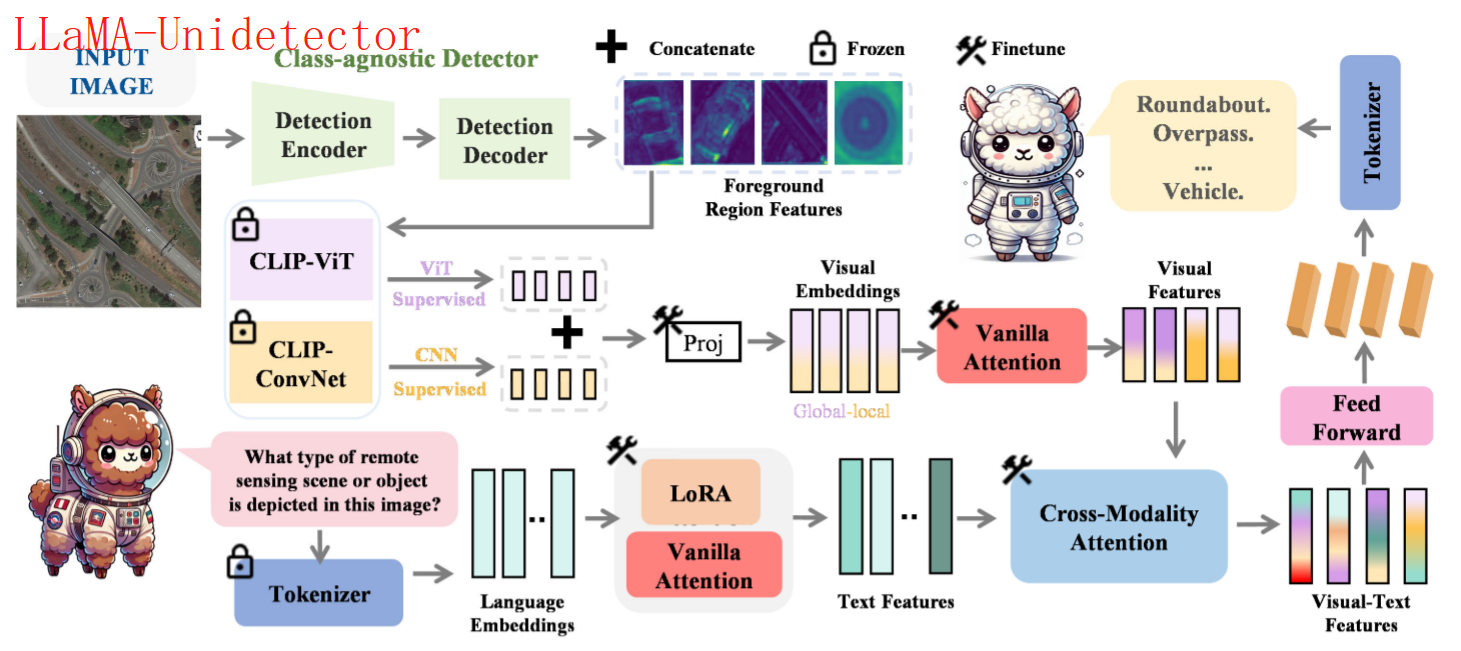}};
        \node[mybox, right=of sm] (smimg) {\includegraphics[width=2.5cm]{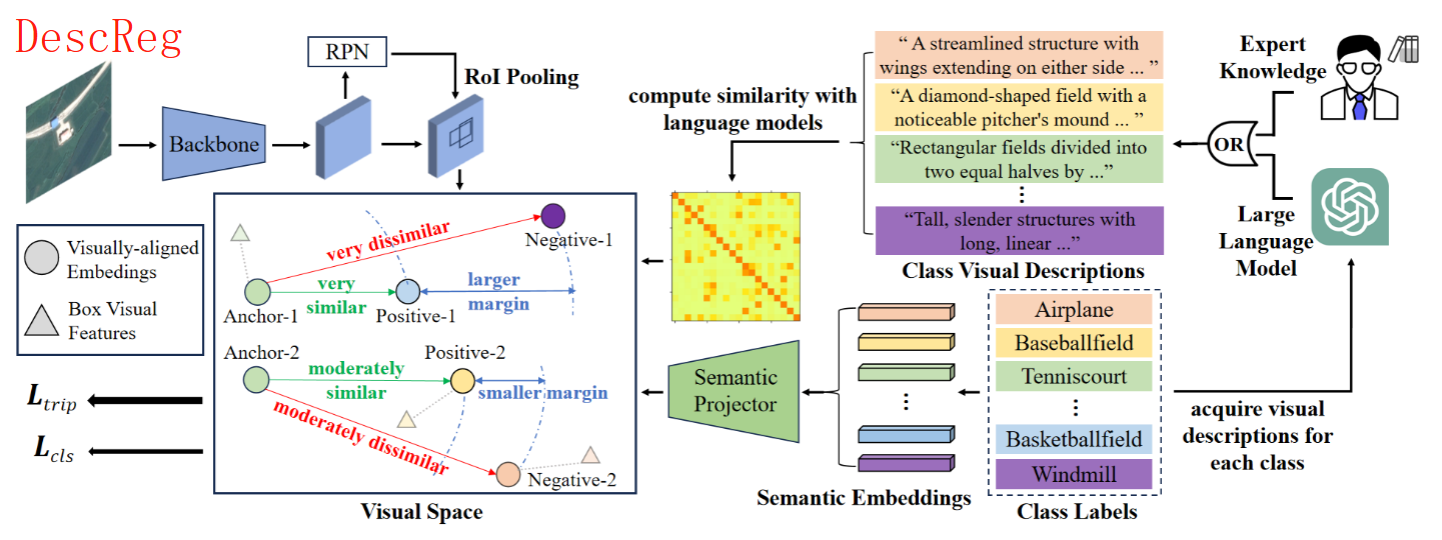}};
        
        \coordinate (fork) at ($(root.east) + (1.1cm, 0)$); 
        \draw (root.east) -- (fork);
        \draw[myarrow] (fork) |- (so.west);
        \draw[myarrow] (fork) |- (kd.west);
        \draw[myarrow] (fork) |- (di.west);
        \draw[myarrow] (fork) |- (er.west);
        \draw[myarrow] (so.east) |- (sf.west);
        \draw[myarrow] (kd.east) |- (wk.west);
        \draw[myarrow] (di.east) |- (sl.west);
        \draw[myarrow] (er.east) |- (sm.west);
        \draw[myarrow] (sf.east) -- (sfimg.west);
        \draw[myarrow] (wk.east) -- (wkimg.west);
        \draw[myarrow] (sl.east) -- (slimg.west);
        \draw[myarrow] (sm.east) -- (smimg.west);
        \end{tikzpicture}%
    }
    \caption{Several typical open-vocabulary object detection methods and their representative works in UAV imagery.}
    \label{fig:typical methods}
\end{figure}

\subsection{CLIP-driven Integration Methods}

This paradigm represents the mainstream approach for building high-performance OVOD detectors. As shown in Figure \ref{fig:tech})(b), the core idea is to tightly couple the vision and language modalities within a single trainable network. Instead of treating classification as a separate step, these models reformulate object detection as a vision language grounding task, where the goal is to localize image regions that correspond to given text prompts. This deep fusion allows the model to learn a rich and shared embedding space, leading to superior detection accuracy.
Within this category, we observe a divergence in design philosophy, primarily centered on balancing performance, efficiency, and universality.

LAE-DINO~\cite{pan2025locate} embodies the data-driven, performance-first approach. Its authors argue that the primary obstacle for aerial OVOD is the "data domain gap." Their main contribution is the construction of LAE-1M, the first million-instance, large-scale OVOD dataset for remote sensing, created using a semi-automatic LAE-Label Engine. Built upon this massive dataset, they propose LAE-DINO, an enhanced version of the powerful Grounding DINO detector. To handle the unprecedented vocabulary size of LAE-1M, they introduce Dynamic Vocabulary Construction (DVC), which samples a smaller and relevant vocabulary for each training batch. To better capture the context of aerial scenes, they designed Visual-Guided Text Prompt Learning (VisGT), a module that enriches text features with scene-level visual information. LAE-DINO establishes a new state-of-the-art in raw performance, demonstrating the 
immense power of training a foundation model on large-scale, in-domain data.

In stark contrast, OVA-Det~\cite{wei2024ova} prioritizes efficiency and real-time performance, critical requirements for onboard UAV processing. Built on the lightweight RT-DETR~\cite{zhao2024detrsbeatyolosrealtime} architecture, OVA-Det introduces several clever, low-overhead modules to achieve image-text collaboration. The core innovations are the Text-Guided Feature Enhancement (TG-FE) and Text-Guided Query Enhancement (TG-QE). These modules use class embeddings as "clues" to guide the encoder and decoder, respectively. They employ cross-attention and a sigmoid gating mechanism to enhance class-relevant features while suppressing background interference, a common issue in cluttered aerial views. This design allows OVA-Det to achieve remarkable zero-shot performance (especially in recall) while operating at an impressive 36 FPS, making it one of the first truly real-time aerial OVOD methods.

Bridging the gap between performance and practicality, OpenRSD~\cite{huang2025openrsd} aims to be a universal and flexible framework. It is designed to be a "Swiss Army knife" for aerial detection. Its key innovation is a dual-head, multi-task architecture. It features a lightweight Alignment Head for fast inference and a more complex Fusion Head for high-precision detection, allowing users to choose the right balance for their application. Critically, OpenRSD is the first framework in this context to explicitly support both horizontal (HBB) and oriented (OBB) bounding boxes, a crucial feature for accurately localizing objects in aerial imagery. It also supports both text and image prompts, further enhancing its flexibility. To maximize generalization, it employs a sophisticated multi-stage training pipeline, including pre-training, fine-tuning, and a carefully designed self-training stage that uses the model's own predictions to create pseudo-labels and refine its knowledge across different datasets.



LLaMA-Unidetector~\cite{LLaMA-Unidetector} takes a skillful approach by decoupling the complex OVOD task into two simpler, distinct stages: object localization and category recognition. Its framework consists of two independent stages. In the first stage, a class-agnostic detector is trained to perform only one task: to identify all potential objects in an image, distinguishing them from the background, without any knowledge of their specific categories. This detector is optimized purely for localization accuracy and generalization. In the second stage, the foreground object regions proposed by the first stage are passed to TerraOV-LLM, a specialized MLLM built upon LLaMA~\cite{llama} and fine-tuned on TerraVQA, a large-scale visual question-answering dataset for remote sensing created by the authors. This MLLM then performs the recognition task, inferring the category of each object based on its visual features and a text prompt. The main advantage of this decoupled approach is its ability to tap into the unparalleled semantic understanding and generalization capabilities of MLLM. This allows LLaMA-Unidetector to achieve outstanding zero-shot performance and even recognize objects with a high degree of contextual or abstract reasoning. The modular design also offers great flexibility, as the localization and recognition components can be upgraded independently. However, this two-stage process introduces significant latency, making it unsuitable for real-time applications. It is also susceptible to compounded errors: if the class-agnostic detector fails to propose a region in the first stage, the MLLM will never have the chance to recognize it, regardless of its power.


DescReg~\cite{zang2024descReg} is a more theoretical and fundamental work that does not propose a new detector architecture but instead focuses on enhancing the quality of the learned visual-semantic embedding space. It identifies a core challenge in aerial Zero Shot Detection (ZSD): the weak semantic-visual correlation. Unlike in natural images, an object's semantic label in aerial imagery may not correlate well with its visual appearance. To solve this, DescReg proposes to regularize the learning process with explicit visual descriptions. For each class, a text description of its typical visual appearance is provided. These descriptions are used to compute a visual similarity matrix between all classes. The core innovation is a novel similarity-aware, adaptive triplet loss. This loss function forces the learned visual feature space to conform to the structure of the pre-defined visual similarity matrix. It ensures that objects described as visually similar are closer in the embedding space than objects described as visually dissimilar, regardless of their semantic labels.

The primary strength of DescReg is its theoretical elegance and high efficiency. It addresses a fundamental problem with a lightweight, mathematically sound solution. As a regularization method, it is highly versatile and can be integrated into various ZSD/OVOD frameworks to boost their performance. However, its effectiveness is dependent on the quality of the visual descriptions provided. It also focuses primarily on the classification aspect and does not directly address the challenge of improving novel object proposal recall in the first place.

\subsection{Summary and Comparison}

\begin{table}
\centering
\caption{Comparison on DIOR and DOTA V1.0 datasets.}
\label{tab: comparison}
\begin{tabular}{l|ccc|ccc}
\toprule
\textbf{Model} & \multicolumn{3}{c|}{\textbf{DIOR}} & \multicolumn{3}{c}{\textbf{DOTA v1.0}} \\
 & Base mAP & Novel mAP & HM & Base mAP & Novel mAP & HM \\
\midrule
\multicolumn{7}{c}{\textbf{mAP-based Evaluation}} \\
\midrule
DescReg & 68.7 & 7.9 & 14.2 & 68.7 & 4.7 & 8.8 \\
CastDet & 51.3 & 24.3 & 33.0 & 60.6 & 36.0 & 45.1 \\
OVA-DETR & 79.6 & 26.1 & 39.3 & 75.5 & 23.7 & 36.1 \\
\midrule
\multicolumn{7}{c}{\textbf{$\mathrm{AP}_{50}$-based Evaluation}} \\
\midrule
LAE-DINO & \multicolumn{3}{c|}{85.5} & \multicolumn{3}{c}{--} \\
OPEN-RSD & \multicolumn{3}{c|}{--} & \multicolumn{3}{c}{77.7} \\
LLaMA-Unidetector & \multicolumn{3}{c|}{51.38} & \multicolumn{3}{c}{50.22} \\
\bottomrule
\end{tabular}
\end{table}

The previous text provides a comparative summary of the two proposed categories of aerial OVOD methods. It highlights their core principles, representative models, and analyzes their strengths and challenges, with a specific focus on their suitability for UAV applications. The analysis reveals a clear spectrum of solutions, ranging from data-efficient semi-supervised methods ideal for low-resource scenarios, to high-performance end-to-end models for accuracy-critical tasks, to highly efficient models for real time deployment, and powerful but slow MLLM-based systems for deep semantic analysis. This diverse landscape underscores the richness of the field and points toward a future where hybrid approaches may combine the best of these paradigms.

Table \ref{tab: comparison} presents a concise performance comparison of various state-of-the-art models on the DIOR and DOTA V1.0 datasets. The evaluation is stratified into two distinct protocols: an mAP-based assessment for few shot or open vocabulary settings and an $\mathrm{AP}_{50}$-based assessment for general detection performance.

In the mAP-based evaluation, we analyze the models' ability to detect both base and novel classes, using the Harmonic Mean (HM) to gauge the balance. A clear trend emerges: while DescReg achieves a high Base mAP of 68.7 on both datasets, its performance on novel classes is critically low, leading to poor HM scores of 14.2 and 8.8, respectively. This indicates significant overfitting to the base categories. In contrast, CastDet and OVA-DETR show a much better trade-off. On DIOR, OVA-DETR leads with the highest HM of 39.3, driven by a strong Base mAP of 79.6. However, on the more challenging DOTA V1.0 dataset, CastDet demonstrates superior generalization to novel classes (36.0 Novel mAP), achieving the best HM of 45.1. DescReg and CastDet were contemporaneous works, while OVA-DETR emerged slightly later. Comparing the two contemporaneous works, DescReg and CastDet, we observe that pseudo-labeling-based algorithms generally achieve superior performance on novel classes. Even when compared to the later-proposed OVA-DETR, CastDet remains highly competitive on novel classes and demonstrates superior performance on the DOTA v1.0 dataset. However, pseudo-labeling-based methods tend to exhibit weaker performance on base classes. We attribute this observation to the inherent characteristics of pseudo-labeling mechanisms. Once pseudo-labels are generated, the model treats novel and base classes similarly during fine-tuning. However, since novel classes receive more focused optimization during this process, the model gradually develops a bias toward novel categories. This bias enhances detection performance for novel classes while simultaneously diverting some of the model's capacity away from base classes, ultimately resulting in comparatively weaker performance on base categories. It should be noted that while pseudo-labeling-based methods demonstrate measurable advantages in novel class detection during evaluation, this performance characteristic does not fully meet the expectations for open-world scenarios.

In the $\mathrm{AP}_{50}$-based evaluation, which measures overall detection accuracy at an IoU threshold of 0.5, we report results for a different set of methods. It is important to note that these scores are not directly comparable to the mAP-based results due to the differing evaluation criteria. LAE-DINO exhibits strong performance on DIOR with an AP50 of 85.5. Similarly, OPEN-RSD achieves a competitive score of 77.7 on DOTA V1.0. LLaMA-Unidetector, evaluated on both datasets, records scores of 51.38 and 50.22, respectively.


\section{Datasets and Evaluation Metrics}
\label{sec:dataset}

A strong and consistent evaluation system is crucial for tracking progress in any research area ~\cite{nilsen2020making,venable2016feds}. For OVOD research, this system depends on two key elements: good datasets that provide realistic challenges, and proper metrics that can fully measure a model's ability to recognize both known and new objects. This section gives a complete review of the main datasets and metrics used in OVOD research, with special attention to how well they work for drone images and where they fall short.

\subsection{General OVOD Datasets}
The foundations of modern OVOD research were laid using large-scale, general-purpose object detection datasets originally designed for closed-set scenarios~\cite{lin2015coco,gupta2019lvisdatasetlargevocabulary,wang2023v3detvastvocabularyvisual,yao2023evaluategeneralizationdetectionbenchmark}. To adapt them for the open-vocabulary task, a standardized protocol of partitioning classes into base and novel sets was established. This split is crucial as it simulates a realistic scenario where a model is trained on a limited set of annotated categories but is expected to operate in an open world with unseen objects.

Base Classes provide the core visual-semantic grounding for the model. The model learns to associate specific visual features with textual class embeddings using the provided bounding box annotations. The primary goal during this phase is to learn a robust and generalizable alignment in a shared embedding space, rather than simply memorizing the base classes;
Novel Classes represent the "open world." The model must detect these objects at test time without ever having seen an annotated example during training. Success on novel classes is the ultimate measure of a model's generalization capability, demonstrating its ability to transfer knowledge from the base classes to new, unseen concepts purely through semantic understanding of their class names.

The COCO dataset~\cite{lin2015coco} is arguably the most influential benchmark in object detection and has become the primary testbed for OVOD. It contains 80 object categories from everyday scenes. For OVOD evaluation, a common protocol is to partition these 80 classes into 65 base classes and 15 novel classes. An alternative, more challenging split designates 48 classes as base and 17 as novel (Removed 15 categories in the WordNet hierarchy that do not have synonym sets). During training, models have access to both images and bounding box annotations for the base classes only. At inference time, the model's performance is evaluated on its ability to detect all 80 classes, with a particular focus on the 15 or 17 novel classes whose textual names are the only supervisory signal provided.

LVIS~\cite{gupta2019lvisdatasetlargevocabulary} was designed to address the long-tail distribution of objects in the real world, featuring over 1200 categories. This characteristic makes it an excellent benchmark for open-vocabulary learning, as it naturally contains a large set of rare classes that can be designated as 'novel'. The class vocabulary of LVIS is often categorized into frequent, common, and rare classes. In the OVOD setting, the frequent and common classes are typically used as the base set, while the large set of rare classes serves as the novel set. This setup rigorously tests a model's ability to generalize from a well-represented base to a sparsely represented, diverse set of unseen categories.

While foundational, these general-purpose datasets, captured primarily from a ground-level perspective, do not fully encapsulate the unique challenges of aerial scenes. The significant domain gap, characterized by top-down viewpoints, vast scale variations, complex backgrounds, and arbitrary object orientations, requires the use of specialized aerial datasets.

\begin{table}[t] 
\caption{Statistics of relevant dataset. The upper section covers typical ground-view detection datasets. The middle section focuses on remote sensing aerial detection datasets, while the lower section is dedicated to datasets specifically designed for open-vocabulary aerial object detection.}
\label{tab: dataset}
\begin{tabular}{l l c c c c}
\toprule
\textbf{Domain} & \textbf{Dataset}	 & \textbf{Class} & \textbf{Image}  & \textbf{Instance}         & \textbf{Annotation way} \\ 
\midrule
\multirow{2}{*}{\centering General scene} 
& COCO~\cite{lin2015coco}             & 80     & 328K   & 2.5M         & -  \\
& LVIS~\cite{gupta2019lvisdatasetlargevocabulary}  & 1200     & 164K   & 2.2M         & -  \\
\midrule
\multirow{20}{*}{\centering Remote Sensing}
& UCAS-AOD~\cite{ucas-aod}   & 2     & 2420   & 14596  & HBB   \\
& RSOD~\cite{RSOD}             & 4     & 3644   & 22221            & OBB     \\
& HRSC~\cite{liu2017hrsc}          & 19    & 1061   & 2976        & OBB      \\
& \multicolumn{1}{c}{NWPU-VHR~\cite{cheng2014vhr}}      & 10    & 800    & 3651     & HBB      \\
& LEVIR~\cite{levir}            & 3     & 21952  & 11028     & HBB        \\
& DOTA-v1.0~\cite{xia2018dota}        & 15    & 2806   & 188282  & OBB       \\
& HRRSD~\cite{HRRSD}            & 13    & 21761  & 55740    & HBB       \\
& SIMD~\cite{SIMD}             & 15    & 5000   & 45096   & HBB        \\
& DIOR~\cite{li2020dior}             & 20    & 23463  & 190288  & HBB       \\
& DIOR-R~\cite{cheng2022dior-r}         & 20    & 23463  &  192512  & OBB   \\
& DOTA-v2.0~\cite{ding2021DOTA2}    & 18    & 11268  & 1973658 & OBB        \\
& xView~\cite{lam2018xview}    & 60    & 1127   & \textgreater{}1M & HBB   \\
& Visdrone~\cite{bansal2018visdrone}   & 10    & 29040  & 740419  & HBB    \\
& FAIR1M~\cite{sun2022fair1m}  & 37    & 15266  & \textgreater{}1M & OBB     \\
& GLH-Bridge~\cite{li2024glh-bridge}       & 1     & 6000   & 59737 & ALL    \\
& SODA~\cite{soda}      & 9     & 31798  & 1008346      & HBB        \\
& RSVG~\cite{rsvg}           & -     & 4239  & 7933            & HBB      \\
& OPT-RSVG~\cite{opt-rsvg}        & 14    & 25452 & 48952  & HBB      \\
& \multicolumn{1}{c}{DIOR-RSVG~\cite{DIOR-RSVG}}        & 20    & 17402 & 38320  & HBB       \\
\midrule
\multirow{3}{*}{\centering Open vocabulary} 
& MI-OAD~\cite{mi-oad}           & 100   & 163023 & 2M      & HBB      \\
& ORSD+~\cite{huang2025openrsd}     & 200   & 474058 & -   & ALL       \\
& LAE-1M~\cite{pan2025locate}           & 1600  & -      & 1M      & HBB    \\
\bottomrule
\end{tabular}
\end{table}

\subsection{UAV-Specific Object Detection Datasets}
The rapid advancement of UAV technology has led to the creation of numerous high-resolution aerial and satellite imagery datasets for object detection. In the following, we review some of the most prominent datasets in this domain, whose characteristics are summarized in the Table \ref{tab: dataset}

Early and HBB-based datasets: Initial efforts in aerial object detection produced datasets like UCAS-AOD~\cite{ucas-aod}, which focuses on two main categories (car and airplane), and NWPU VHR-10~\cite{cheng2014vhr}, which expands to 10 common object classes. These datasets were instrumental in early research but are limited by their relatively small scale and use of Horizontal Bounding Boxes (HBB). HBBs are often imprecise for aerial objects like ships and airplanes, which are non-axis-aligned, leading to the inclusion of significant background noise within the bounding box. DIOR~\cite{li2020dior} is a much larger scale HBB dataset, with 20 classes and over 190,000 instances, serving as a comprehensive benchmark for HBB-based detection in complex remote sensing scenes.

The Rise of OBB for precise localization: A major leap forward came with the introduction of Oriented Bounding Boxes (OBB). The HRSC~\cite{liu2017hrsc} dataset was a pioneer in this area, providing OBB annotations for ships and highlighting the need for rotation-aware detection. This trend was solidified by the DOTA series~\cite{xia2018dota,ding2021DOTA2}. DOTA-v1.0~\cite{xia2018dota} became a de facto standard, with 15 categories, 2,806 large images, and over 188,000 OBB-annotated instances. Its successor, DOTA-v2.0~\cite{ding2021DOTA2}, further expanded this with 18 categories and a staggering 1.97 million instances, presenting immense challenges in scale, orientation, and aspect ratio.

Towards million-instance and fine-grained recognition: The last few years have seen the emergence of massive-scale datasets. xView~\cite{lam2018xview} contains over 1 million instances across 60 fine-grained categories, although it uses HBB annotations. FAIR1M~\cite{sun2022fair1m} pushed the boundaries further by providing over 1 million instances with high-quality OBB annotations, organized into a hierarchical structure of 5 super-categories and 37 sub-categories. More recently, STAR~\cite{li2025star} continues this trend with a fine-grained hierarchy of 8 super-categories and 48 sub-categories and OBB annotations. These datasets are crucial for training data-hungry models and evaluating fine-grained recognition capabilities. Other specialized datasets like GLH-Bridge~\cite{li2024glh-bridge} focus on a single, challenging category with OBB.

The LAE-1M~\cite{pan2025locate} dataset represents a significant effort to solve the problem of data scarcity in the remote sensing community. It is the first large-scale remote sensing dataset to reach one million labeled objects with a broad category coverage. The construction of LAE-1M is particularly innovative, employing a two-pronged LAE-Label Engine: 
Fine-grained Data (LAE-FOD): For existing, human-labeled datasets, the engine unifies them by performing image slicing, format alignment, and strategic sampling. This creates a high-quality, fine-grained object detection subset.
Coarse-grained Data (LAE-COD): To leverage the vast amount of unlabeled aerial imagery, the engine uses a semi-automated pipeline. It first employs a segmentation model to generate region proposals from unlabeled images. Then, a powerful large VLMs is used to assign categorical labels to these regions in a zero-shot manner. Finally, rule-based filtering cleans the auto-generated labels.
By combining both fine-grained and coarse-grained data, LAE-1M provides an unprecedented scale and diversity, containing around 1600 unique vocabulary terms. This makes it an ideal pretraining corpus for developing foundational OVOD models for the aerial domain.

The ORSD+~\cite{huang2025openrsd} dataset, proposed alongside the OpenRSD framework, is another large-scale training dataset constructed to enhance cross-domain generalization. It comprises over 470,000 images spanning 200 categories. The key innovation of ORSD+ lies in its multi-stage construction and training pipeline:
Data Aggregation: It begins by integrating numerous existing public datasets, both labeled and unlabeled, which creates an initial and heterogeneous data pool.
ORSD+ is explicitly designed to train a universal remote sensing detector that can handle both HBB and OBB tasks and perform robustly across datasets it was not explicitly fine-tuned on, making it highly relevant for open-world scenarios.

The MI-OAD~\cite{mi-oad} dataset directly tackles the most significant limitation of prior works: the lack of rich, descriptive textual annotations. While datasets like DOTA~\cite{ding2021DOTA2} use simple word-level labels, MI-OAD pioneers the creation of a massive dataset with sentence-level descriptions for objects. It is by far the largest dataset of its kind, containing 163,023 images and 2 million image-caption pairs, which is approximately 40 times larger than any previous remote sensing visual grounding dataset. 
Its "Open-Source Word-to-Sentence (OS-W2S) Label Engine" is designed to generate rich annotations:
Instead of just categories, the engine uses a powerful VLMs to generate detailed captions for each object. These captions describe not only the object's class but also its attributes, its relative position to other objects, and its absolute position within the image.
It provides annotations at three distinct levels: vocabulary-level, phrase-level, and sentence-level.
Meanwhile, it breaks the one-to-one correspondence between a caption and a single object. A single descriptive caption can be associated with multiple instances in the image, mimicking real-world language use.
MI-OAD is the first benchmark truly designed to evaluate fine-grained, open vocabulary aerial detection, moving beyond simple category names to complex, descriptive natural language prompts.

Despite these incredible new dataset-building efforts, a standardized benchmark specifically partitioned for Open-Vocabulary Object Detection in the UAV aerial domain is still in its infancy. While LAE-1M~\cite{pan2025locate} and MI-OAD~\cite{mi-oad} propose their own evaluation splits, a community-wide consensus has yet to form. Therefore, the community urgently needs to develop a dedicated UAV-OVOD benchmark that will facilitate progress, achieve fair comparisons, and guide research to address the practical challenges of open world perception in aerial scenes. In addition, from the dataset summarized in Table \ref{tab: dataset}, we can see a paradigm shift in dataset creation for the aerial domain:
\begin{itemize}
    \item First, we observe a dramatic increase in category diversity. While conventional remote sensing datasets typically contain several dozen classes at most, open-vocabulary datasets like ORSD+~\cite{huang2025openrsd} and LAE-1M~\cite{pan2025locate} push this to hundreds or even over a thousand categories, reflecting a significant step towards capturing real-world semantic diversity.
    \item Second, this leap in scale is enabled by a fundamental change in data creation methodology. Rather than relying solely on manual annotation, each of these pioneering datasets is constructed using custom designed annotation engines. These engines are critically dependent on the powerful zero shot and text generation capabilities of modern VLMs not only to scale up instance numbers but also to to increase the semantic richness of the supervision, thus laying the groundwork for a new generation of foundation models for Earth observation.
\end{itemize}



\subsection{Evaluation Metrics}
To comprehensively evaluate an OVOD model, metrics must capture its performance on both the familiar base classes and the unseen novel classes~\cite{mAP,recall}. Standard object detection metrics are adapted for this dual-objective evaluation. Several widely used metrics are:
\begin{itemize}
    \item \textbf{Average Precision}. The cornerstone of object detection evaluation is Average Precision (AP). AP provides a single-figure measure that summarizes the quality of a detector by considering both its ability to correctly classify objects (Precision) and its ability to find all relevant objects (Recall). To understand AP, we must first define its components: precision and recall. For a given object class, precision measures the fraction of correct predictions among all predictions made for that class. Recall measures the fraction of correct predictions among all ground-truth instances of that class. These two metrics can be formulated as:
    \begin{align}
    \text{Precision} &= \frac{\text{TP}}{\text{TP} + \text{FP}} \\
    \text{Recall} &= \frac{\text{TP}}{\text{TP} + \text{FN}}
    \end{align}
    Where True Positives (TP) are correctly detected objects, False Positives (FP) are incorrect detections, and False Negatives (FN) are missed ground truth objects. A detection is typically considered a True Positive if its Intersection over Union (IoU) with a ground-truth box is above a certain threshold. The IoU threshold is a critical hyperparameter that defines the required strictness of spatial accuracy. A widely used threshold is 0.5 (or 50\%). Metrics reported with this threshold are often denoted as $AP_{\text{50}}$ or mAP@0.5.
    \item \textbf{Precision-Recall}. Precision-Recall (PR) Curve is an ideal detector which measures precision and  recall simultaneously. However, there is often a trade-off: to increase recall (find more objects), a model may lower its confidence threshold, which can lead to more false positives and thus lower precision. The PR curve visualizes this trade-off by plotting precision against recall for various confidence thresholds.
    \item \textbf{Calculating Average Precision}. AP is conceptually defined as the area under the PR curve. To create a more stable and representative metric, modern evaluation protocols~\cite{pascal,10coco} employ an interpolation method. The precision at any given recall level is set to the maximum precision achieved at any recall level greater than or equal to it. This creates a monotonically decreasing PR curve, and the AP is the area under this interpolated curve.
    \item \textbf{Mean Average Precision}. Mean average precision(mAP) is the primary metric for object detection. It is calculated separately for the base and novel class sets. $mAP_{\text{base}}$: This metric is computed over the set of base classes. It quantifies the model's ability to retain its detection performance on the classes it was explicitly trained on. A high $mAP_{\text{base}}$ indicates that the model has not suffered from "catastrophic forgetting" while learning to generalize. $mAP_{\text{novel}}$: This is the most critical metric for OVOD. It is computed exclusively over the set of novel classes. It directly measures the model's generalization power, its ability to locate and classify objects it has never seen before. A high $mAP_{\text{novel}}$ signifies effective knowledge transfer from the seen to the unseen.
    \item \textbf{Harmonic Mean}. To provide a single, balanced score that reflects a model's overall OVOD capability, the harmonic mean (HM) of the base and novel mAP scores is widely used. It is calculated as:
    \begin{equation}
    H = \frac{2 \cdot \text{mAP}_{\text{base}} \cdot \text{mAP}_{\text{novel}}}{\text{mAP}_{\text{base}} + \text{mAP}_{\text{novel}}}
    \end{equation}
    The HM is more informative than a simple arithmetic mean because it heavily penalizes models that exhibit a large disparity between base and novel class performance. For instance, a model that achieves a very high $mAP_{\text{base}}$ but a near-zero $mAP_{\text{novel}}$ will receive a very low H-score. This metric thus encourages the development of models that achieve a strong balance between retaining knowledge of seen classes and successfully generalizing to new ones, which is the central goal of open-vocabulary object detection.
    
\end{itemize}

\section{Challenges and Open Issues}
\label{sec:challenge}

The combination of open vocabulary object detection (OVOD) and drone technology offers exciting possibilities for real world applications. However, this integration also faces several key challenges. Aerial images have special characteristics, and current OVOD methods still have limitations, which create difficulties that need to be solved to make this technology fully effective~\cite{survey1,zhu2021detectiontrackingmeetdrones,survey3}. This chapter examines these challenges in detail and discusses open research questions. We divide them into two main categories: (1) challenges caused by the nature of drone-captured aerial scenes, and (2) challenges related to current OVOD techniques and how they align text and visual data.

\subsection{Limitations of UAV scenarios}

\subsubsection{The Domain Gap}

A major challenge in applying open-vocabulary object detection (OVOD) to drone imagery stems from the significant differences between standard visual-language model (VLM) training data and actual UAV-captured footage. Advanced VLMs like CLIP~\cite{clip} and similar models form the foundation of most OVOD systems, achieving their impressive zero-shot recognition through pretraining on billions of web images paired with text descriptions. However, these training images primarily feature ground-level perspectives showing objects in familiar side or frontal views, typically well-framed and centered against common everyday backgrounds that match human visual experience.

In contrast, drone-captured imagery presents fundamentally different characteristics. The aerial perspective introduces substantial visual distortion - when viewed from above, cars appear as simple rectangles while people become small dots, losing the distinctive visual features models rely on for recognition. Objects also exhibit extreme size variations depending on altitude, requiring models to identify the same object whether it fills most of the frame at low altitude or appears just a few pixels wide when flying high. Additionally, the complex, cluttered backgrounds typical of urban or natural aerial scenes make object detection particularly challenging, as targets must be distinguished from dense environmental features~\cite{view,scale,du2018unmannedaerialvehiclebenchmark}.

This domain gap significantly impacts model performance. The semantic connections learned during training, such as associating "passenger bus" with side-view features like length and windows - become unreliable when applied to overhead views where these characteristics aren't visible. Consequently, models often fail to recognize objects that are obvious to human observers, even when the visual evidence is clear in the aerial imagery. This fundamental mismatch between ground-level training data and aerial operational conditions remains a critical obstacle for effective UAV-based object detection systems.

Here are some Open Issues:
\begin{itemize}
\item   \textbf{Domain Adaptation for VLMs: }How can we adapt pretrained VLMs to the aerial domain without compromising their open-vocabulary capabilities? Fine-tuning on limited aerial data risks overfitting and catastrophic forgetting of the vast knowledge learned during pre-training. Research into parameter-efficient fine-tuning (PEFT) techniques~\cite{hu2021lora,li-liang-2021-prefix}, such as adapters~\cite{hu2023llmadaptersadapterfamilyparameterefficient} or prompt tuning~\cite{lester2021powerscaleparameterefficientprompt}, is a promising direction.
\item   \textbf{Synthetic Data Generation: }Can we leverage simulation engines to generate large-scale, photorealistic aerial datasets with precise, multi-perspective annotations? This could help bridge the domain gap by exposing the VLM to top-down views during a secondary pre-training or fine-tuning stage.
\item	\textbf{Viewpoint-Invariant Feature Learning: }Developing novel network architectures or training strategies that encourage the learning of viewpoint-invariant object representations is a key research goal. This might involve contrastive learning objectives that pull representations of the same object from different viewpoints closer in the embedding space.
\end{itemize}

\subsubsection{The Small Object Detection}
The detection of small objects is a long-standing problem in computer vision, and it is particularly acute in UAV imagery due to high flight altitudes~\cite{rssmall,wang2022normalizedgaussianwassersteindistance}. In the context of OVOD, this challenge is amplified. A small object, often defined as occupying a very small number of pixels, presents a dual dilemma:

Information Scarcity in Visual Features: Standard hierarchical vision backbones progressively down sample the input image to build semantic representations. This process, while effective for large objects, can cause the features of small objects to diminish or vanish entirely in deeper, more semantically rich layers. The resulting feature vector for a small object proposal is often weak, noisy, and lacks the discriminative information necessary for robust recognition.

Difficulty in Visual-Semantic Alignment: The core mechanism of OVOD is to align region-specific visual features with text-embedding vectors. When the visual feature vector is sparse and non-descriptive due to the object's small size, achieving a meaningful and confident alignment with a rich textual description becomes exceptionally difficult. The model struggles to differentiate a few blurry pixels corresponding to a "life raft" from background noise or other small, irrelevant objects, especially when both have low-quality feature representations.

While traditional small object detection methods employ specialized techniques like feature pyramid networks (FPNs)~\cite{fpn}, high-resolution feature fusion, and context-aware modules, integrating these seamlessly into the OVOD framework is non-trivial. The challenge lies not just in enhancing small object features but in ensuring that these enhanced features are compatible with the VLM's pretrained embedding space.

Here are some Open Issues:
\begin{itemize}
\item	\textbf{High-Resolution OVOD Architectures: }Designing OVOD models that can process high-resolution imagery efficiently and maintain fine-grained spatial detail throughout the network is crucial. This may involve exploring novel multi-scale backbones or attention mechanisms tailored for small object feature preservation.
\item	\textbf{Context-Enhanced Alignment: }For small objects, surrounding context often provides critical clues. How can we design models that explicitly leverage contextual information (e.g., a "boat" is on "water") to aid the visual-semantic alignment for the small object itself? This could involve graph-based reasoning or attention mechanisms that correlate object proposals with their environmental context.
\item	\textbf{Super-Resolution as a Pre-processing Step: }Investigating the use of generative super-resolution techniques to "hallucinate" details for small objects before feature extraction could be a viable, albeit computationally expensive, strategy. The challenge is to ensure the generated details are faithful and do not introduce misleading artifacts.
\end{itemize}

\subsubsection{The Fine-Grained Recognition}
UAV are often deployed for tasks that require not just detecting object categories but distinguishing between visually similar sub-categories, a task known as fine-grained recognition. For example, in traffic monitoring, it is essential to differentiate between a "truck," a "bus," a "van," an "SUV," and a "sedan." From a high-altitude, top-down perspective, these distinctions are incredibly subtle.
This challenge is magnified in the OVOD setting for two primary reasons:

Loss of Distinguishing Features: The key visual cues that differentiate fine-grained categories are often subtle and localized (e.g., the length-to-width ratio, the presence of a cargo bed on a truck, the specific shape of the hood). From an aerial view, these features can be obscured, distorted, or simply too small to be resolved. All vehicles may appear as similarly colored rectangles, making visual differentiation based on intrinsic features nearly impossible.

High Demands on Language Specificity: OVOD relies on the text prompt to guide detection. To perform fine-grained recognition, the model must understand the subtle semantic differences between prompts like "truck" and "bus" and link them to the minimal available visual evidence. The VLM may have learned these distinctions from ground-level images where a "bus" has a long row of windows and a "truck" has a separate cab and trailer. When these features are absent in the aerial view, the model's ability to ground the specific text prompt correctly is severely compromised.

This challenge pushes the limits of a VLM's ability to generalize. The model must infer the correct category from indirect cues like relative size, location (e.g., on a highway vs. a city street), or roof features, which may not have been explicitly encoded in its original training.

Here are some Open Issues:
\begin{itemize}

\item	\textbf{Hierarchical and Attribute-Based Prompting: }Instead of a single class name, can we use more descriptive, attribute-based prompts (e.g., "a long vehicle with a flat roof," "a small four-wheeled vehicle")? Developing methods that can parse and reason about such compositional queries is a key research area. Hierarchical prompting (e.g., querying for "vehicle" first, then refining to "truck" or "bus") could also be a viable strategy.

\item	\textbf{Injecting Domain-Specific Knowledge: }Can we explicitly inject domain knowledge into the model? For example, a knowledge graph could inform the model that "buses are typically longer than SUVs" or that "fishing boats are found near coastlines." Integrating this symbolic knowledge with the model's learned representations could significantly improve fine-grained accuracy.
\item	\textbf{Few-Shot Fine-Grained Learning: }In many applications, an operator may want to find a new, specific type of object. This requires the model to learn a new fine-grained category from just one or a few visual examples, a challenging few-shot learning problem within the OVOD context.
\end{itemize}

\subsubsection{The trade-off between efficiency and performance}
A significant practical barrier to the widespread deployment of OVOD on UAV is the classic trade-off between model performance and computational efficiency. High-performing OVOD models, particularly those based on large backbones like ViT~\cite{vit}, are computationally demanding and have a large memory footprint.

Computational Cost: The attention mechanisms in Transformers~\cite{vaswani2023attentionneed} have a quadratic complexity with respect to the number of image patches, making high-resolution processing very expensive. The sheer number of parameters in models like ViT~\cite{vit} requires substantial GPU resources for inference.

Resource-Constrained Platforms: UAVs are fundamentally resource-constrained edge devices. They operate on limited battery power, and their onboard processing units have a fraction of the computational power and memory of a server-grade GPU.

This creates a dilemma: deploying a large, powerful model is often infeasible due to hardware and power limitations, while deploying a smaller, more efficient model may lead to an unacceptable drop in detection accuracy, especially for the challenging aerial scenarios described above. Real-time performance is often a strict requirement for applications like autonomous navigation or immediate threat detection, further tightening the constraints.

Here are some Open Issues:
\begin{itemize}
\item	\textbf{Model Compression for OVOD: }Research is urgently needed in adapting model compression techniques for OVOD models. This includes:
Quantization: Reducing the precision of model weights to decrease memory usage and accelerate computation on compatible hardware.
Pruning: Removing redundant weights or network structures to create a smaller, sparser model.
Knowledge Distillation: Training a small, efficient "student" model to mimic the output of a large, high-performance "teacher" VLMs. The key challenge is how to effectively distill the rich and open-vocabulary knowledge.

\item	\textbf{Efficient OVOD Architectures: }There is a need for the design of novel, lightweight network architectures specifically for edge-based OVOD. This might involve hybrid CNN-Transformer models or architectures that are optimized for the specific hardware accelerators found on UAVs.

\item	\textbf{Cloud-Edge Collaborative Systems: }An alternative approach is a hybrid system where the UAV performs lightweight, on-board pre-processing (e.g., detecting potential regions of interest) and transmits only relevant data to a more powerful ground station or cloud server for full OVOD analysis. The primary challenge here is managing communication latency and bandwidth.
\end{itemize}

\subsection{Limitations of text prompting methods}
\subsubsection{Ambiguity and Robustness of Text Prompts}
The "open vocabulary" capability of OVOD is both its greatest strength and a potential source of fragility. The performance of the system is highly dependent on the quality and specificity of the user-provided text prompts. This introduces challenges of ambiguity and robustness.

Semantic Ambiguity: The choice of words can have a significant impact on detection results. For instance, the prompts "car," "automobile," and "vehicle" may seem synonymous to a human, but they can produce different results. "Vehicle" is broad and might correctly identify cars but also trigger false positives on buses and trucks. "Car" might be too specific and fail to detect SUVs or minivans if the VLM's internal concept of "car" is biased towards sedans. This "prompt engineering" is currently more of an art than a science.

Varying Levels of Abstraction: Users may wish to query for abstract concepts rather than concrete objects. For example, in a disaster response scenario, a relevant query might be "signs of damage," "a dangerous situation," or "a gathering crowd." These prompts do not correspond to a well-defined object category. They require a higher level of scene understanding and reasoning that current OVOD models, which are primarily trained for object-level recognition, are not equipped to handle. Grounding such abstract concepts in visual evidence is a frontier research problem.

Lack of Robustness: Models can be brittle. A slight rephrasing of a prompt or the inclusion of descriptive adjectives can sometimes lead to unpredictable changes in performance. Furthermore, most models lack the ability to understand negation or complex compositional queries involving spatial relationships.

Here are some Open Issues:
\begin{itemize}

\item	\textbf{Automated Prompt Engineering and Refinement: }Can we develop methods that automatically generate or refine text prompts to be optimal for a given task or dataset? This could involve learning a mapping from a user's high-level intent to a set of effective, low-level prompts.

\item	\textbf{Learning from Multiple Prompts: }Instead of relying on a single prompt, models could be designed to leverage a set of synonymous or related prompts to produce more robust and reliable detections.

\item	\textbf{Abstract and Compositional Reasoning: }A major leap forward would be the development of OVOD systems that can handle abstract queries by decomposing them into recognizable visual components. For example, a "dangerous situation" on a highway might be decomposed into "overturned car," "traffic jam," and "emergency vehicles.
\end{itemize}

\subsubsection{The Lack of Benchmarks}

A final, overarching challenge that impedes progress in the entire field of UAV-OVOD is the critical lack of standardized benchmarks. The development and evaluation of new methods are currently fragmented and difficult to compare.
Most existing UAV datasets~\cite{lin2015coco,xia2018dota,ding2021DOTA2,li2020dior} were created for traditional, closed-set object detection. While they provide high-quality aerial imagery and bounding box annotations, their class vocabularies are fixed and relatively small. Researchers wishing to evaluate OVOD models on this data must manually define their own "base" and "novel" class splits, leading to inconsistent evaluation protocols. Furthermore, these datasets lack the rich, descriptive, and hierarchical text labels needed to fully test the capabilities of VLMs.

A professional benchmark needs to meet the following requirements
Enable Fair Comparison: Provide a common ground with standardized training, validation, and test sets, along with pre-defined base and novel vocabularies, to allow for the direct and fair comparison of different methods.
Drive Progress on Key Challenges: The benchmark should be explicitly designed to include challenging scenarios that target the problems outlined in this chapter: a wide distribution of object scales (especially small objects), fine-grained categories with subtle differences, and diverse viewpoints and backgrounds.
Standardize Evaluation Metrics: Define clear and comprehensive evaluation metrics. This should include not only mAP on base and novel classes but also metrics for evaluating performance on hierarchical and descriptive queries, robustness to prompt variations, and computational efficiency.

In conclusion, while the fusion of OVOD and UAV technology holds immense promise, the path to robust, real-world deployment is fraught with challenges. Addressing the domain gap, solving the small object and fine-grained recognition puzzles, balancing performance with efficiency, improving prompt robustness, and establishing standardized benchmarks are the key open issues that will define the research agenda in this exciting field for years to come.

\section{Future Perspectives and Directions}
\label{sec:future}
The integration of OVOD into UAV aerial scenes, while promising, is still in its nascent stages. The challenges detailed in the previous chapter not only highlight the current limitations but also illuminate a clear path forward for future research. To transition this technology from a laboratory concept to a robust, deployable real-world system, the community must focus on several key research thrusts. This chapter outlines our vision for the future of UAV-OVOD, presenting six pivotal directions that we believe will shape the landscape of this powerful field.

\subsection{Domain Adaptation for UAV-OVOD}
The performance of current OVOD models is fundamentally constrained by the domain gap between ground-level web data and top-down aerial imagery. Bridging this gap is arguably the most critical step towards unlocking reliable performance. Future research must move beyond the naive application of off-the-shelf VLMs and focus on creating aerial-aware models that retain their open-vocabulary prowess.

A primary direction is the exploration of PEFT techniques. Methods such as Adapters~\cite{hu2023llmadaptersadapterfamilyparameterefficient}, which insert small, trainable modules between the frozen layers of a pretrained model, or LoRA~\cite{hu2021lora}, which injects trainable low-rank matrices into Transformer layers, offer a compelling solution. These approaches allow the model to learn aerial-specific features using a small amount of UAV data while keeping the vast majority of the VLM's parameters frozen. This strategy mitigates the risk of "catastrophic forgetting," thereby preserving the rich semantic knowledge learned from web-scale data.

Another promising avenue is secondary pretraining on a curated, mid-scale corpus of aerial imagery. This involves taking a general-purpose VLM and continuing its pre-training on a dataset composed of existing aerial imagery and, crucially, large-scale synthetic aerial data. Advanced simulation platforms can generate photorealistic aerial scenes with perfect, automatic annotations, providing a cost-effective way to expose the model to a massive volume of top-down visual concepts. The research challenge lies in optimizing this secondary pretraining phase to maximize domain adaptation without incurring prohibitive computational costs.

\subsection{Lightweight and Efficient OVOD Models}
The practical utility of UAV-OVOD is contingent upon its ability to run in real-time on resource-constrained onboard processors. The "performance vs. efficiency" trade-off must be addressed through dedicated research into lightweight and efficient OVOD architectures.

Model compression techniques will be paramount. This includes:
\begin{itemize}
    \item Quantization: Systematically reducing the numerical precision of model weights and activations. Post-training quantization and quantization-aware training tailored for OVOD models need to be investigated to minimize the accuracy loss.
    
    \item Knowledge Distillation: This is a particularly powerful paradigm for OVOD. A large, high-performance "teacher" model can be used to train a small, efficient "student" model. The key research question is what knowledge to distill. Beyond simply matching the final detection outputs, the student could be trained to mimic the teacher's intermediate region-text feature alignments, thereby learning the rich cross-modal relationships that enable open-vocabulary recognition.
    
    \item Network Pruning and Architecture Search: Exploring structured pruning to remove entire filters or attention heads, and employing Neural Architecture Search (NAS) to discover novel, hardware-aware network designs that are inherently optimized for the computational patterns of edge GPUs.

\end{itemize}
The ultimate goal is to create a family of OVOD models that offer a flexible trade-off, allowing operators to choose a model that meets the specific latency and accuracy requirements of their mission.

\subsection{Multi-Modal Data Fusion}
UAVs are often equipped with a suite of sensors beyond standard RGB cameras, such as thermal infrared (IR), multispectral, or even LiDAR sensors. The future of UAV-OVOD lies in moving beyond the RGB-Text paradigm to a more holistic multi-modal fusion framework. This will enable robust perception in challenging conditions where RGB data is ambiguous or unavailable, such as at night, in fog, or through smoke.

Imagine a search-and-rescue mission at night. A query for "person" using an RGB camera would likely fail. However, by fusing data from a thermal camera, the system could be prompted to find a "heat source" or "a warm object with a human-like shape." This requires developing novel fusion architectures. Instead of simple early or late fusion, sophisticated cross-modal attention mechanisms could allow features from one modality to dynamically inform and enhance features from another. For instance, thermal features could guide the attention of the RGB-processing stream towards salient regions, and vice-versa. The text embedding would then be aligned with this fused, multi-modal feature representation, creating a powerful system for all-weather detection. Research in this area will need to address challenges in cross-modal alignment, data synchronization, and training with heterogeneous data sources.

\subsection{Interactive and Conversational Detection}
The current interaction model for OVOD—a single, static text prompt—is limiting. The future will see a shift towards interactive and conversational systems that allow for a natural, multi-turn dialogue between the human operator and the UAV. This paradigm, inspired by advances in Embodied AI and Vision-Language Navigation, would enable complex, context-aware instructions.
Consider the following interaction:


\begin{figure}
\centering
\includegraphics[width= 0.5\textwidth]{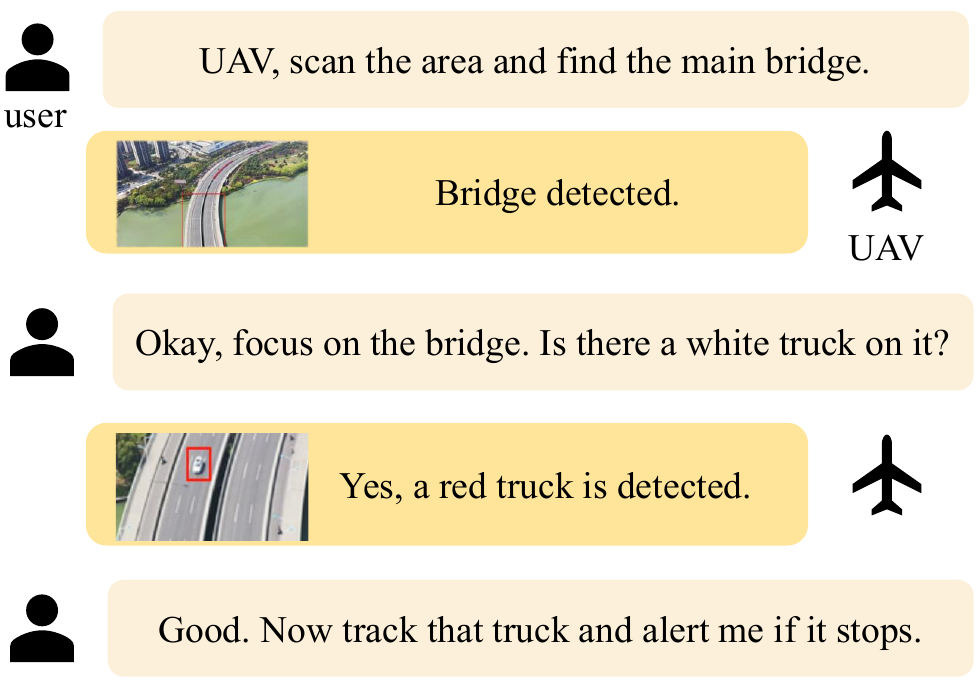}
\caption{Interaction between human operators and drones.\label{fig6}}
\end{figure} 
This level of interaction requires more than just detection. It necessitates models that can handle:
Dialogue History and State Tracking: Maintaining the context of the conversation;
Coreference Resolution: Understanding that "it" in the second prompt refers to the "bridge" from the first;
Task Grounding: Translating natural language commands ("track that truck") into specific actions for the perception and control modules.
This research direction points towards a powerful synergy between LLMs for understanding conversational intent and OVOD models for grounding that intent in the visual world.

\subsection{Building Large-Scale UAV-OVOD Benchmarks}
Progress in any data-driven field is catalyzed by high-quality public benchmarks. The lack of a standard benchmark is a major bottleneck for UAV-OVOD research. A concerted effort from both academia and industry is imperative to build a large-scale, richly-annotated UAV-OVOD benchmark.

This future benchmark should possess several key characteristics:
\begin{itemize}

    \item Scale and Diversity: It must contain tens of thousands of images from diverse geographical locations, altitudes, times of day, and weather conditions.
    \item Expansive and Hierarchical Vocabulary: The vocabulary should encompass hundreds or even thousands of object classes, from common categories to rare instances and fine-grained sub-categories.
    \item Rich Annotations: Crucially, annotations must go beyond simple bounding boxes; A tight bounding box and/or a segmentation mask; Attribute labels; Multiple free-form textual descriptions, capturing the object's appearance and context.

\end{itemize}
Such a benchmark would not only enable fair and reproducible comparisons but also drive research towards solving the core challenges of fine-grained recognition, context-based reasoning, and descriptive language grounding.

\subsection{Integration with Downstream Tasks}
Ultimately, OVOD is an enabling perception technology, not an end in itself. Its true value will be realized through its seamless integration into a broader ecosystem of intelligent UAV tasks, evolving from object detection to holistic situational awareness.
Future research will focus on creating unified models or pipelines that leverage OVOD as a foundational component for more complex capabilities:
\begin{itemize}
    \item Open-Vocabulary Tracking (OVT): Extending the "detect-by-description" capability to "track-by-description." An operator could initiate tracking by simply describing the target, and the system would maintain a persistent track of that specific object across time and viewpoint changes.
    \item Open-Vocabulary Segmentation: Moving beyond bounding boxes to provide pixel-level masks for any described object or region. This is vital for tasks like precise area measurement or damage assessment.
    \item UAV-based Visual Question Answering (VQA) and Scene Captioning: Enabling an operator to ask complex questions about the aerial scene or receive dense, language-based summaries of the dynamic environment.
\end{itemize}

By integrating these capabilities, we move towards a future where a UAV can autonomously perceive, understand, and describe its surroundings in human-like terms, transforming it from a simple remote sensor into a truly intelligent partner for a wide range of critical applications.

\reftitle{References}


\bibliography{reference}

\isChicagoStyle{%

}{}

\isAPAStyle{%

}{}

%


\PublishersNote{}
\end{document}